\documentclass[runningheads]{llncs}
\usepackage{latexsym}

\usepackage{times}
\usepackage{soul}
\usepackage{url}
\usepackage[hidelinks]{hyperref}
\usepackage[utf8]{inputenc}
\usepackage{graphicx}
\usepackage{booktabs}
\usepackage{algorithm}
\usepackage{algorithmic}
\usepackage{url,ifthen}
\usepackage{amssymb}

\begin{document}

\title{Stabilized Nested Rollout Policy Adaptation}
\author{Tristan Cazenave\inst{1} \and Jean-Baptiste Sevestre\inst{2} \and Matthieu Toulemont\inst{3}}

\authorrunning{T. Cazenave et al.}

\institute{LAMSADE, Université Paris-Dauphine, PSL, CNRS, France \email{Tristan.Cazenave@dauphine.psl.eu}
\and InstaDeep \email{jb.sevestre@instadeep.com} \and \email{matthieu.toulemont@ponts.org}}

\maketitle
                                                                              
\begin{abstract}
Nested Rollout Policy Adaptation (NRPA) is a Monte Carlo search algorithm for single player games. In this paper we propose to modify NRPA in order to improve the stability of the algorithm. Experiments show it improves the algorithm for different application domains: SameGame, Traveling Salesman with Time Windows and Expression Discovery.
\end{abstract}

\section{Introduction}

Monte Carlo Tree Search (MCTS) has been successfully applied to many
games and problems \cite{BrownePWLCRTPSC2012}.

Nested Monte Carlo Search (NMCS) \cite{CazenaveIJCAI09} is
an algorithm that works well for puzzles and optimization problems. It biases its playouts using lower level playouts. At level zero NMCS adopts a uniform random playout policy. Online learning of playout strategies combined with NMCS has given good results on optimization problems
\cite{RimmelEvo11}. Other applications of NMCS include Single Player General Game Playing \cite{Mehat2010}, Cooperative Pathfinding \cite{Bouzy13}, Software testing \cite{PouldingF14}, heuristic Model-Checking \cite{PouldingF15}, the Pancake problem \cite{Bouzy16}, Games \cite{CazenaveSST16} and the RNA inverse folding problem \cite{portela2018unexpectedly}.

Online learning of a playout policy in the context of nested searches
has been further developed for puzzles and optimization with Nested
Rollout Policy Adaptation (NRPA) \cite{Rosin2011}. NRPA has found new
world records in Morpion Solitaire and crosswords puzzles. Stefan
Edelkamp and co-workers have applied the NRPA algorithm to multiple
problems. They have optimized the algorithm for the Traveling Salesman
with Time Windows (TSPTW) problem
\cite{cazenave2012tsptw,edelkamp2013algorithm}. Other applications
deal with 3D Packing with Object Orientation \cite{edelkamp2014monte},
the physical traveling salesman problem \cite{edelkamp2014solving},
the Multiple Sequence Alignment problem \cite{edelkamp2015monte} or
Logistics \cite{edelkamp2016monte}. The principle of NRPA is to adapt
the playout policy so as to learn the best sequence of moves found so
far at each level. Unfortunately, this mechanism only samples each policy once at the lowest level which may lead to a misclassification of a good policy (one that improves the best score) as a bad one. To solve this issue, we propose a simple, yet effective modification of the NRPA Algorithm, which we name Stabilized NRPA.By sampling each policy multiple times at the lowest level we show that this new NRPA is stabilized and converges faster. 

We now give the outline of the paper. The second section describes NRPA. The third section explains Stabilized NRPA. The fourth section describes the problems used for the experiments. The fifth section gives experimental results for these problems. The sixth section outlines further work and the last section concludes.

\section{NRPA}

Nested Policy Rollout Adaptation is an algorithm introduced by Chris Rosin \cite{Rosin2011} that achieves state-of-the-art performance on problems such as Morpion Solitaire.

\begin{figure*}[ht]
    \centering
        \hspace*{-0.5cm}\includegraphics[scale=0.40]{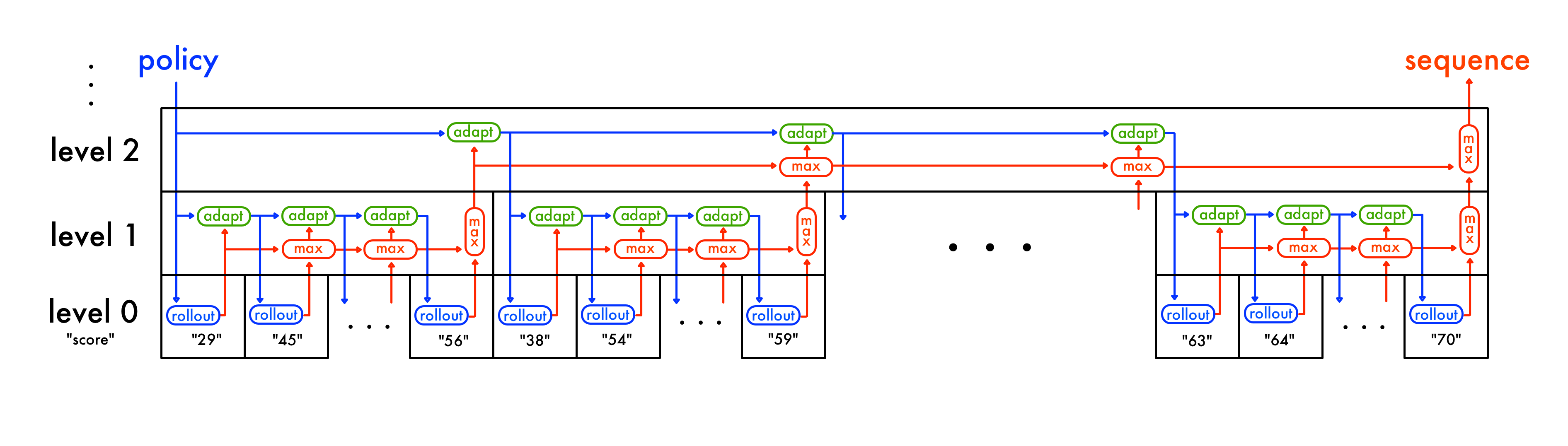}
    \caption{NRPA scheme}
    \label{fig:nrpa}
\end{figure*}

This algorithm has two major components : An adaptive rollout policy, and a nested structure, shown in Figure \ref{fig:nrpa}.

The adaptive rollout policy is a policy parameterized by weights on each action. During the playout phase, action is sampled according to this weights. The Playout Algorithm is given in algorithm \ref{PLAYOUT}. It uses Gibbs sampling, each move is associated to a weight. A move is coded as an integer that gives the index of its weight in the policy array of floats. The algorithm starts with initializing the sequence of moves that it will play (line 2). Then it performs a loop until it reaches a terminal states (lines 3-6). At each step of the playout it calculates the sum of all the exponentials of the weights of the possible moves (lines 7-10) and chooses a move proportionally to its probability given by the softmax function (line 11). Then it plays the chosen move and adds it to the sequence of moves (lines 12-13).

Then, the policy is adapted on the best current sequence found, by increasing the weight of the best actions. The Adapt Algorithm is given in algorithm \ref{ADAPT}.For all the states of the sequence passed as a parameter it adds $\alpha$ to the weight of the move of the sequence (lines 3-5). Then it reduces all the moves proportionally to $\alpha$ times the probability of playing the move so as to keep a sum of all probabilities equal to one (lines 6-12).

The nested structure was introduced by Tristan Cazenave \cite{CazenaveIJCAI09}. 
This method helps the algorithm to converge towards better and better sequences.
In NRPA, each nested level takes as input a policy, and returns a sequence.
Inside the level, the algorithm makes many recursive calls to lower levels, providing weights, getting sequences and adapting the weights on those sequences. In the end, the algorithm returns the best sequence found in that level. At the lowest level, the algorithm simply makes a rollout.

The NRPA algorithm is given in algorithm \ref{NRPA}. At level zero it simply performs a playout (lines 2-3). At greater levels it performs N iterations and for each iteration it calls itself recursively to get a score and a sequence (lines 4-7). If it finds a new best sequence for the level it keeps it as the best sequence (lines 8-11). Then it adapts the policy using the best sequence found so far at the current level (line 12).

NRPA balances exploitation by adapting the probabilities of playing moves toward the best sequence of the level, and exploration by using Gibbs sampling at the lowest level. It is a general algorithm that has proven to work well for many optimization problems.

\begin{algorithm}
\begin{algorithmic}[1]
\STATE{Playout ($state$, $policy$)}
\begin{ALC@g}
\STATE{$sequence$ $\leftarrow$ []}
\WHILE{true}
\IF{$state$ is terminal}
\RETURN{(score ($state$), $sequence$)}
\ENDIF
\STATE{$z$ $\leftarrow$ 0.0}
\FOR{$m$ in possible moves for $state$}
\STATE{$z$ $\leftarrow$ $z$ + exp ($policy$ [code($m$)])}
\ENDFOR
\STATE{choose a $move$ with probability $\frac{exp (policy [code(move)])}{z}$}
\STATE{$state$ $\leftarrow$ play ($state$, $move$)}
\STATE{$sequence$ $\leftarrow$ $sequence$ + $move$}
\ENDWHILE
\end{ALC@g}
\end{algorithmic}
\caption{\label{PLAYOUT}The Playout algorithm}
\end{algorithm}

\begin{algorithm}
\begin{algorithmic}[1]
\STATE{Adapt ($policy$, $sequence$)}
\begin{ALC@g}
\STATE{$polp \leftarrow$ $policy$}
\STATE{$state \leftarrow$ $root$}
\FOR{$move$ in $sequence$}
\STATE{$polp$ [code($move$)] $\leftarrow$ $polp$ [code($move$)] + $\alpha$}
\STATE{$z$ $\leftarrow$ 0.0}
\FOR{$m$ in possible moves for $state$}
\STATE{$z$ $\leftarrow$ $z$ + exp ($policy$ [code($m$)])}
\ENDFOR
\FOR{$m$ in possible moves for $state$}
\STATE{$polp$ [code($m$)] $\leftarrow$ $polp$ [code($m$)] - $\alpha * \frac{exp (policy [code(m)])}{z}$}
\ENDFOR
\STATE{$state$ $\leftarrow$ play ($state$, $move$)}
\ENDFOR
\STATE{$policy$ $\leftarrow$ $polp$}
\RETURN{$policy$}
\end{ALC@g}
\end{algorithmic}
\caption{\label{ADAPT}The Adapt algorithm}
\end{algorithm}

\begin{algorithm}
\begin{algorithmic}[1]
\STATE{NRPA ($level$, $policy$)}
\begin{ALC@g}
\IF{level == 0}
\RETURN{playout (root, $policy$)}
\ELSE
\STATE{$bestScore$ $\leftarrow$ $-\infty$}
\FOR{N iterations}
\STATE{(result,new) $\leftarrow$ NRPA($level-1$, $policy$)}
\IF{result $\geq$ bestScore}
\STATE{bestScore $\leftarrow$ result}
\STATE{seq $\leftarrow$ new}
\ENDIF
\STATE{policy $\leftarrow$ Adapt (policy, seq)}
\ENDFOR
\RETURN{(bestScore, seq)}
\ENDIF
\end{ALC@g}
\end{algorithmic}
\caption{\label{NRPA}The NRPA algorithm.}
\end{algorithm}

\section{Stabilized NRPA}

In this section we explain Stabilized NRPA and its potential for being parallelized.

\begin{figure*}[ht]
    \centering
         \hspace*{-0.5cm}\includegraphics[scale=0.40]{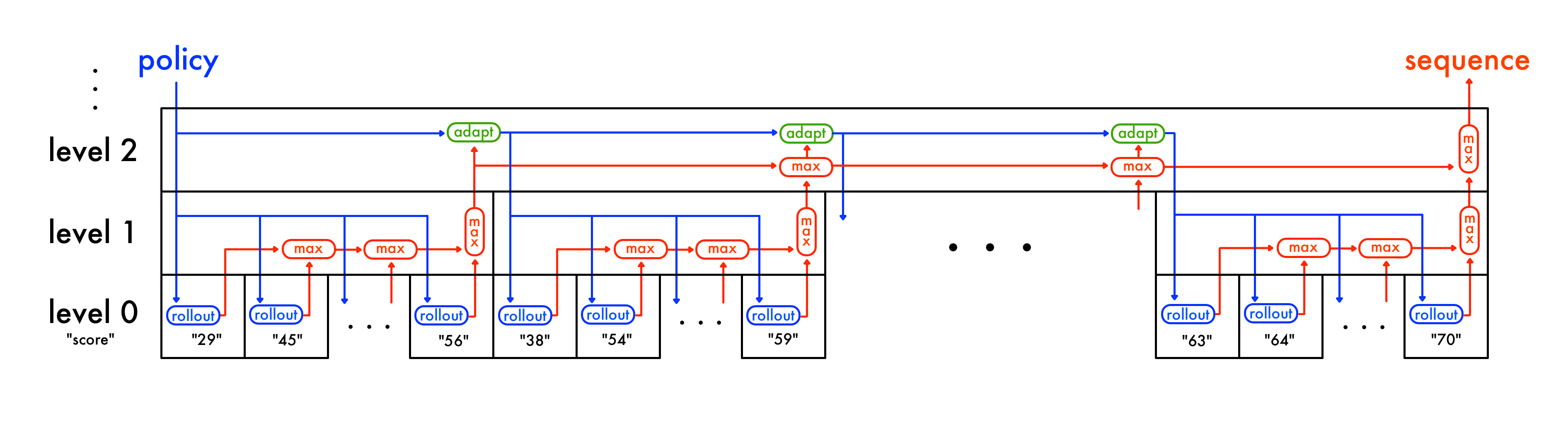}
    \caption{Stabilized NRPA scheme. Level 1 is replaced by an evaluation level}
    \label{fig:snrpa}
\end{figure*}

\subsection{Better Convergence of NRPA}

In NRPA algorithm, an evaluation problem may occur.

Imagine that we have a policy that has good performance, but unfortunately the sequence generated by this policy at level 0 is bad (i.e. the sequence has a bad score comparing to the usual policy performance).
This sequence is up to level 1 and is ignored since it is worse than the best sequence of level 1. The policy is adapted on the best sequence of level 1, pushing slightly the next rollouts toward the best sequence of level 1, making the policy more deterministic, making it less exploratory and less likely to find a new best sequence. This bad behavior could be propagated to the upper level, for the same reasons.

The problem is even worse when this situation occurs at the beginning of a nested level since there is not yet a best sequence. In this case the policy is adapted directly on this bad sequence, pushing the rollouts towards bad sequences, which perturbs the rollouts of the entire nested level.

To prevent this problem, an idea is simply to generate not only $1$ sequence according to a given policy, but $P$ sequences, in order to get a better evaluation of the performance of this policy. The algorithm does not adapt to the best sequence until $P$ sequence have been played. And the best sequence returned is the best sequence over those $P$ sequences.

We note that doing so stabilizes the convergence of NRPA. Rollouts are less often pushed to bad sequences, making entire nested level less perturbed, and so making each nested level useful for the search efficiency, leading also to faster convergence.

In our experiments, we have replaced classic level 1 by an evaluation level leading to Figure \ref{fig:snrpa}, that aims to better evaluate the policy, and to return the best sequence found by this policy. We can see in figure \ref{fig:snrpa} that multiple level zero calls are performed before doing the adapt in green whereas in figure \ref{fig:nrpa} the green adapt function is called after every level zero call.

The number of evaluation is parameterized by the $P$ parameter and the number of playouts at the lowest level of SNRPA is $P$ times greater than the number of playout at the lowest level of NRPA.

Note that for a fixed number of playouts, the Stabilized NRPA makes less updates comparing to NRPA, making it faster.
Note further that Stabilized NRPA is a generalization of NRPA, since SNRPA(1) is NRPA.

Stabilized NRPA is given in algorithm \ref{StabilizedNRPA}. It follows the same pattern as NRPA. Lines 2-3 and lines 14-25 are the same as in NRPA. They correspond to level zero and to levels strictly greater than one. The difference lies in level one (lines 4-13). At level one there is an additional loop from 1 to $P$ that gets the best sequence out of $P$ playouts.

\begin{algorithm}
\begin{algorithmic}[1]
\STATE{StabilizedNRPA ($level$, $policy$)}
\begin{ALC@g}
\IF{level == 0}
\RETURN{playout (root, $policy$)}
\ELSIF{level == 1}
\STATE{$bestScore$ $\leftarrow$ $-\infty$}
\FOR{1, \dots, P}
\STATE{(result,new) $\leftarrow$ StabilizedNRPA($level-1$, $policy$)}
\IF{result $\geq$ bestScore}
\STATE{bestScore $\leftarrow$ result}
\STATE{seq $\leftarrow$ new}
\ENDIF
\ENDFOR
\RETURN{(bestScore, seq)}

\ELSE
\STATE{$bestScore$ $\leftarrow$ $-\infty$}
\FOR{1, \dots, N}
\STATE{(result,new) $\leftarrow$ StabilizedNRPA($level-1$, $policy$)}
\IF{result $\ge$ bestScore}
\STATE{bestScore $\leftarrow$ result}
\STATE{seq $\leftarrow$ new}
\ENDIF
\STATE{policy $\leftarrow$ Adapt (policy, seq)}
\ENDFOR
\RETURN{(bestScore, seq)}
\ENDIF
\end{ALC@g}
\end{algorithmic}
\caption{\label{StabilizedNRPA}The Stabilized NRPA algorithm.}
\end{algorithm}

\subsection{Parallelization}

Parallelizing NMCS was done in \cite{CazenaveJ09}. Parallelizing NRPA on a cluster is easily done using root parallelization when distributing among the different computers and using leaf parallelization on each multiple cores computer \cite{NegrevergneC17}. More recently Andrzej Nagorko efficiently parallelized NRPA while not changing its global behavior \cite{Nagorko19}.

Stabilized NRPA is well fitted for leaf parallelization as the P playouts can be done in parallel.

\section{Problems Used for the Experiments }

In this section we present the three problems used for the experiments. The Maximum problem where the goal is to find a mathematical expression that evaluates as great as possible. The TSPTW problem that finds short paths to visit as set of cities with time constraints. The SameGame problem, a popular puzzle.

\subsection{The Maximum Problem}

Nested Monte Carlo Search can be used for the optimization of mathematical expressions \cite{CazenaveExpression10,Cazenave13Discovery,CazenaveH15}. For some problems it gives better results than alternative algorithms such as UCT \cite{Kocsis2006} or Genetic Programming \cite{koza1994genetic}.

The Maximum problem \cite{langdon1997analysis} consists in finding an expression that results in the maximum possible number given some limit on the size of the expression. In the experiment limit was on the depth of the corresponding tree and the available atoms were +, * and 0.5. In our experiments we fixed a limit on the number of atoms of the generated expression, not on the depth of the tree and the available atoms are +, * and 1.0.

We applied NRPA to the Maximum Problem. It is the first time NRPA is applied to Expression Discovery.

\begin {figure} [htb]
\begin{center}  
  \begin{tabular}{ccc}
    \includegraphics [width=3.0cm]  {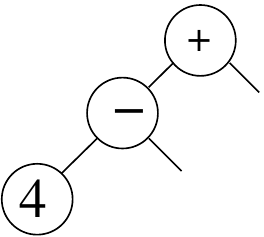} &
    \includegraphics [width=3.0cm]  {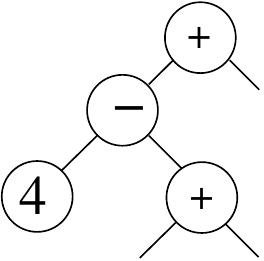} & ~~~~~~~~~~~~~
    \includegraphics [width=0.5cm]  {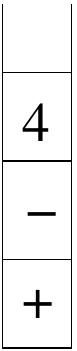}
  \end{tabular}
  {\caption {A partial tree and the corresponding stack.}
  \label{tree}}
\end{center}
\end {figure}

Figure \ref{tree} gives an example of how an expression is built using a playout. The left tree corresponds to the stack of atoms below the tree. The stack defines a tree and in order to fill the tree new atoms are pushed on top of the stack. For example pushing the '+' atom on the stack gives the tree on the right. When the maximum number of nodes + leaves is reached for a stack only terminal atoms (atoms that do not have children) are pushed onto the stack enforcing the number of nodes of the generated expression to be below the defined maximum.

\subsection{TSPTW}

In the Traveling Salesman Problem with Time Windows (TSPTW) an agent has to visit N cities at predefined periods of times while minimizing the total tour cost. NRPA has been successfully applied to TSPTW \cite{cazenave2012tsptw,edelkamp2013algorithm}.

The Hamiltonian Path problem is a subproblem of the TSP, so TSPTW and most other TSP variants are computationally hard. No algorithm polynomial in the number of cities is expected.

The TSPTW is much harder than the TSP, different algorithms have to be used for solving this problem and NRPA had state of the art results on standard benchmarks.

Following the formulation of  \cite{cazenave2012tsptw}, the TSPTW can be defined as follow.
Let $G$ be an undirected complete graph.
$G=(N,A)$, where $N=0,1,\ldots,n$ corresponds to a set of nodes and $A=N \times N$ corresponds to the set of edges between the nodes. The node $0$ corresponds to the depot. Each city is represented by the $n$ other nodes.
A cost function $c: A \rightarrow R$ is given and represents the distance between two cities. A solution to this problem is a sequence of nodes $P=(p_0,p_1,\ldots,p_n)$ where $p_0=0$ and $(p_1,\ldots,p_n)$ is a permutation of $[1,N]$. Set $p_{n+1}=0$ (the path must finish at the depot), then the goal is to minimize the function defined in Equation \ref{eq:lecout}.
\begin{equation}
cost(P) = \sum_{k=0}^{n}c(a_{p_k},a_{p_{k+1}})\label{eq:lecout}
\end{equation}

As said previously, the TSPTW version is more difficult because each city $i$ has to be visited in a time interval $[e_i,l_i]$. This means that a city $i$ has to be visited before $l_i$. It is possible to visit a cite before $e_i$, but in that case, the new departure time becomes $e_i$. Consequently, this case may be dangerous as it generates a penalty. Formally, if $r_{p_k}$ is the real arrival time at node $p_k$, then the departure time $d_{p_k}$ from this node is $d_{p_k}=max(r_{p_k},e_{p_k})$.\\

In the TSPTW, the function to minimize is the same as for the TSP (Equation \ref{eq:lecout}), but a set of constraint is added and must be satisfied. Let us define $\Omega (P)$ as the number of violated windows constraints by tour (P).\\
Two constraints are defined.
The first constraint is to check that the arrival time is lower than the fixed time. Formally,
$$
\forall p_k, r_{p_k}<l_{p_k}.
$$

The second constraint is the minimization of the time lost by waiting at a city. Formally,
$$
r_{p_{k+1}}=\max(r_{p_k},e_{p_k})+c(a_{p_k,p_{k+1}}).
$$

In NRPA paths with violated constraints can be generated. As presented in \cite{RimmelEvo11} , a new score $Tcost(p)$ of a path $p$ can be defined as follow:
$$
Tcost(p) = cost(p) + 10^6 * \Omega(p),
$$
with, as defined previously, $cost(p)$ the cost of the path $p$ and $\Omega(p)$ the number of violated constraints.
$10^6$ is a constant chosen high enough so that the algorithm first optimizes the constraints.

The problem we use to experiment with the TSPTW problem is the most difficult problem from the set of \cite{potvin1996vehicle}.
 
\subsection{SameGame}

\begin{figure}
\centering
\includegraphics[scale=0.8]{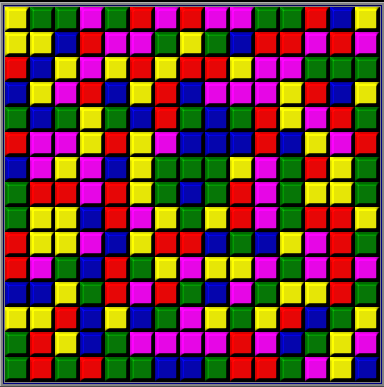}
\caption{First problem of the SameGame standard suite}
\label{SameGamePb1}
\vspace{-6mm}
\end{figure}

In SameGame the goal is to score as much as possible removing connected components of the same color. An example of a SameGame board is given in figure \ref{SameGamePb1}. The score for removing $n$ tiles is $(n-2)^2$. If the board is completely cleared there is a bonus of 1000.

When applying Monte Carlo Search to SameGame it is beneficial to use selective search \cite{cazenave2016selective} in order to eliminate moves that are often bad. For example it is important to remove the tiles of the dominant color all at once in order to score a lot with this move. The Tabu color playout strategy achieves this by forbidding moves of the dominant color when they do not clear all the tiles of the dominant color in one move. We sometimes allow moves of size two for the dominant color beside the Tabu color strategy as advised in \cite{cazenave2016selective}.

The best published results for SameGame come from a parallel implementation of NRPA \cite{NegrevergneC17}.

Figure \ref{SameGamePb1} gives the first problem of the standard SameGame suite. This is the one we used in our experiments.

\section{Experimental Results}

\begin{table*}[ht]
  \centering
  \caption{Results for the Maximum problem (scale $\times$ 1 000 000).}
  \label{tableMaximum}
  \begin{tabular}{rrrrrrrrrrr}
 Time       &  NRPA & SNRPA(2) & SNRPA(3) & SNRPA(4) & SNRPA(10) \\
 \\
0.01 & 1 & 1 & \textbf{2} & \textbf{2} &  1  \\
0.02 & 3 & 24 & 68 & 138 & 25 \\
0.04 & 16 & 108 & 326 & 934 & \textbf{5086} \\
0.08 & 150 & 1341 & 2092 & 5971 & \textbf{21745} \\
0.16 & 3475 & 15844 & 19874 & 52041 & \textbf{88547} \\
0.32 & 170265 & 534672 & 487983 & \textbf{1147083} & 789547 \\
0.64 & 13803062 & 28885199 & 22863271 & \textbf{36529536} & 12000748 \\
1.28 & 40077774 & 216376610 & 270326701 & 379573875 & 212668695 \\
2.56 & 89668935 & 314740908 & 408327339 & 495249021 & \textbf{708820733}\\
5.12 & 151647343 & 472960557 & 557957691 & 704240083 & \textbf{904642720} \\
10.24 & 345707890 & 712902227 & 856149587 & 938008979 & \textbf{1296603114} \\
20.48 & 852761999 & 1151948749 & 1284225823 & 1359946097 & \textbf{1661398711} \\
40.96 & 1975250628 & 2168737831 & 2221426342 & \textbf{2232301333} & 2128244879 \\
81.92 & 2973605038 & 3276850130 & \textbf{3381032884} & 3321287204 & 3057041220 \\
163.84 & 3336604131 & 3531572024 & 3627351674 & 3621195107 & \textbf{3928494648}
\end{tabular}
\end{table*}

\begin{table}
  \centering
  \caption{Best scores for SameGame}
  \label{tablerecord}
  \begin{tabular}{rrrrrrrrrrrrrr}
Problem & NMCS & SP-MCTS & ~~~~~NRPA & SRNPA(4) & js-games.de \\ 
   &  &  &  &  &  \\
 1  & 3,121 & 2,919 & 3,179 & 3,203 & 3,413 \\
 2  & 3,813 & 3,797 & 3,985 & 3,987 & 4,023 \\
 3  & 3,085 & 3,243 & 3,635 & 3,757 & 4,019 \\
 4  & 3,697 & 3,687 & 3,913 & 4,001 & 4,215 \\
 5  & 4,055 & 4,067 & 4,309 & 4,287 & 4,379 \\
 6  & 4,459 & 4,269 & 4,809 & 4,799 & 4,869 \\
 7  & 2,949 & 2,949 & 2,651 & 2,151 & 3,435 \\
 8  & 3,999 & 4,043 & 3,879 & 4,079 & 4,771 \\
 9  & 4,695 & 4,769 & 4,807 & 4,821 & 5,041 \\
 10 & 3,223 & 3,245 & 2,831 & 3,333 & 3,937 \\
 11 & 3,147 & 3,259 & 3,317 & 3,531 & 3,783 \\
 12 & 3,201 & 3,245 & 3,315 & 3,355 & 3,921 \\
 13 & 3,197 & 3,211 & 3,399 & 3,379 & 3,821 \\
 14 & 2,799 & 2,937 & 3,097 & 3,121 & 3,263 \\
 15 & 3,677 & 3,343 & 3,559 & 3,783 & 4,161 \\
 16 & 4,979 & 5,117 & 5,025 & 5,377 & 5,517 \\
 17 & 4,919 & 4,959 & 5,043 & 5,049 & 5,227 \\
 18 & 5,201 & 5,151 & 5,407 & 5,491 & 5,503 \\
 19 & 4,883 & 4,803 & 5,065 & 5,325 & 5,343 \\
 20 & 4,835 & 4,999 & 4,805 & 5,203 & 5,217 \\
    &  &  &  &  &  \\
Total & 77,934 & 78,012 & 80,030 & 82,032 & 87,858 \\
  \end{tabular}
\end{table}

\begin{table*}[!ht]
  \centering
  \caption{Results for the TSPTW rc204.1 problem}
  \label{tableTSPTW}
  \begin{tabular}{rrrrrrrrrrr}
 Time       &  NRPA & SNRPA(2) & SNRPA(3) & SNRPA(4) & SNRPA(10) \\
 \\
0.01 & -29037026 & \textbf{-28762022} & -29107010 & -29222032 & -29337060 \\
0.02 & -26501832 & -26121858 & -26226870 & -26181904 & -27096936 \\
0.04 & -25276756 & -24221694 & -24056722 & -23596696 & -24031802 \\
0.08 & -23821720 & -22621656 & -22556632 & -22176624 & -21706624 \\
0.16 & -22006640 & -21436606 & -21216568 & -20806566 & -20261500  \\
0.32 & -19521526 & -19441520 & -19481502 & -19086484 & -18821438 \\
0.64 & \textbf{-16416390} & -16536396 & -16536403 & -16536387 & -17166394\\
1.28 & -13966259 & -13636262 & -13466266 & \textbf{-13316265} & -14691306\\
2.56 & -12781221 & -11881189 & -11111173 & -10856164 & -11696195 \\
5.12 & -11301179 & -10556154 & -9866131 & -9406120 & \textbf{-8831112} \\
10.24 & -9351129 & -8816107 & -8166091 & -7866081 & \textbf{-7241065}\\
20.48 & -6591049 & -6631047 & -6166038 & \textbf{-6031033} & -6076040 \\
40.96 & \textbf{-3695987} & -3890987 & -3975989 & -4045989 & -4085994 \\
81.92 & -1825960 & -1560955 & \textbf{-1505955} & -1540954 & -2100962 \\
163.84 & -980946 & -780941 & -580938 & -500938 & \textbf{-385937}
 \end{tabular}
\end{table*}

\begin{table*}[!ht]
  \centering
  \caption{Results for the first problem of SameGame}
  \label{tableSameGame}
  \begin{tabular}{rrrrrrrrrrr}
 Time       &  NRPA & SNRPA(2) & SNRPA(3) & SNRPA(4) & SNRPA(5) & SNRPA(6) & SNRPA(7) & SNRPA(10) \\
 \\
0.01 & 448 & 499 & 483 & 494 & \textbf{504} & 479 & 485 & 464 \\
0.02 & 654 & 685 & 678 & \textbf{701} & 676 & 672 & 660 & 637 \\
0.04 & 809 & 871 & 863 & 896 & \textbf{904} & 867 & 836 & 823 \\
0.08 & 927 & 989 & 1010 & \textbf{1062} & \textbf{1062} & 1045 & 1032 & 1026 \\
0.16 & 1044 & 1091 & 1133 & 1183 & 1172 & 1156 & 1170 & \textbf{1186} \\
0.32 & 1177 & 1214 & 1239 & 1286 & 1286 & 1278 & 1285 & \textbf{1288} \\
0.64 & 1338 & 1370 & 1386 & 1407 & \textbf{1414} & 1396 & 1403 & 1398 \\
1.28 & 1514 & 1548 & 1559 & 1556 & \textbf{1573} & 1544 & 1548 & 1547 \\
2.56 & 1662 & 1739 & 1721 & 1740 & \textbf{1759} & 1713 & 1733 & 1716 \\
5.12 & 1790 & 1859 & 1894 & 1900 & 1913 & \textbf{1917} & 1913 & 1897 \\
10.24 & 1928 & 2046 & 2025 & 2034 & 2068 & \textbf{2080} & 2071 & 2065 \\
20.48 & 2113 & 2228 & 2249 & \textbf{2277} & 2255 & 2271 & 2243 & 2213 \\
40.96 & 2393 & 2518 & 2475 & \textbf{2556} & 2518 & 2513 & 2471 & 2477\\
81.92 & 2642 & 2753 & 2718 & \textbf{2787} & 2761 & 2760 & 2733 & 2700 \\
163.84 & 2838 & 2898 & 2868 & \textbf{2949} & 2940 & 2945 & 2912 & 2917
 \end{tabular}

\end{table*}

In all our experiments we use the average score over 200 runs. For each search time starting at 0.01 seconds and doubling until 163.84 seconds we give the average score reached by Stabilized NRPA within this time. We run a search of level 4 each run, but NRPA does not has the time to finish level 3, especially when running SNRPA(P). SNRPA(P) advances approximately P times less steps than NRPA at level 3 since it spends approximately P times more at level 1. All the experiments use sequential versions of NRPA and Stabilized NRPA.

Table \ref{tableMaximum} gives the evolution for the Maximum problem. The score is the evaluation of the mathematical expression. The first column gives the average scores of standard NRPA. The second column gives the average scores of Stabilized NRPA with $P=2$. The third column gives the average scores of Stabilized NRPA with $P=3$ and so on. We can observe that SNRPA(10) gives the best results. To save place the numbers generated by the expressions have been divided by 1 000 000.

Table \ref{tableTSPTW} gives the results for the rc204.1 TSPTW problem. This is the most difficult problem of the Solomon-Potwin-Bengio TSPTW benchmark. The score is one million times the number of violated constraints plus the tour cost. SNRPA(10) gives again the best results.

Table \ref{tableSameGame} gives the results for the first problem of SameGame. Evaluation improves the performance until a certain limit. Indeed, $P=4$ provides the best results with $P=5$ and $P=6$ yielding close scores.

For the three problems, Stabilized NRPA gives better results than NRPA.

Among the different version of SNRPA, the conclusion differs depending of the problem we consider :

For the Maximum Problem, we note that values as great as 10 for $P$ give the best results. For the TSPTW Problem, we note that for the longest time (163.84s), we go from $-980 946$ for NRPA, to $-385 937$ for SNRPA(10) the best result for the greatest value we have tried for $P$. On the contrary smaller values for $P$ seem appropriate for SameGame with $P=4$ being the best.

Table\ref{tablerecord} gives the scores reached by different algorithms on the standard test set of 20 SameGame problems. We see that SNRPA(4) improves on NRPA at level 4. However SNRPA(4) takes more time when run sequentially since it does four times more playouts as NRPA. Still is does the same number of calls to the adapt function as NRPA. SP-MCTS is a variant of the UCT algorithm applied to single player games, and NMCS is Nested Monte Carlo Search. They both reach smaller overall scores than SNRPA(4). the last column contains the records from the website js-games.de. They were obtained by specialized algorithms and little is known about these algorithms except that some of them use a kind of beam search with specialized evaluation functions.



\section{Conclusion}

Stabilized NRPA is a simple modification of the NRPA algorithm. It consists in periodically playing $P$ playouts at the lowest level before performing the adaptation. It is a generalization of NRPA since Stabilized NRPA with $P=1$ is NRPA. It improves the average scores of NRPA given the same computation time for three different problems: Expression Discovery, TSPTW and SameGame.

\section*{Acknowledgment}

This work was supported in part by the French government under management of Agence Nationale de la Recherche as part of the “Investissements d’avenir” program, reference ANR19-P3IA-0001 (PRAIRIE 3IA Institute).

\bibliographystyle{splncs04}
\bibliography{main}

\begin{thebibliography}{10}
\providecommand{\url}[1]{\texttt{#1}}
\providecommand{\urlprefix}{URL }
\providecommand{\doi}[1]{https://doi.org/#1}

\bibitem{Bouzy13}
Bouzy, B.: Monte-carlo fork search for cooperative path-finding. In: Computer
  Games - Workshop on Computer Games, {CGW} 2013, Held in Conjunction with the
  23rd International Conference on Artificial Intelligence, {IJCAI} 2013,
  Beijing, China, August 3, 2013, Revised Selected Papers. pp. 1--15 (2013)

\bibitem{Bouzy16}
Bouzy, B.: Burnt pancake problem: New lower bounds on the diameter and new
  experimental optimality ratios. In: Proceedings of the Ninth Annual Symposium
  on Combinatorial Search, {SOCS} 2016, Tarrytown, NY, USA, July 6-8, 2016. pp.
  119--120 (2016)

\bibitem{BrownePWLCRTPSC2012}
Browne, C., Powley, E., Whitehouse, D., Lucas, S., Cowling, P., Rohlfshagen,
  P., Tavener, S., Perez, D., Samothrakis, S., Colton, S.: A survey of {M}onte
  {C}arlo tree search methods. {IEEE} Transactions on Computational
  Intelligence and {AI} in Games  \textbf{4}(1),  1--43 (Mar 2012).
  \doi{10.1109/TCIAIG.2012.2186810}

\bibitem{CazenaveIJCAI09}
Cazenave, T.: {Nested Monte-Carlo Search}. In: Boutilier, C. (ed.) IJCAI. pp.
  456--461 (2009)

\bibitem{CazenaveExpression10}
Cazenave, T.: Nested monte-carlo expression discovery. In: {ECAI} 2010 - 19th
  European Conference on Artificial Intelligence, Lisbon, Portugal, August
  16-20, 2010, Proceedings. pp. 1057--1058 (2010).
  \doi{10.3233/978-1-60750-606-5-1057},
  \url{https://doi.org/10.3233/978-1-60750-606-5-1057}

\bibitem{Cazenave13Discovery}
Cazenave, T.: Monte-carlo expression discovery. International Journal on
  Artificial Intelligence Tools  \textbf{22}(1) (2013).
  \doi{10.1142/S0218213012500352},
  \url{https://doi.org/10.1142/S0218213012500352}

\bibitem{cazenave2016selective}
Cazenave, T.: Nested rollout policy adaptation with selective policies. In: CGW
  at IJCAI 2016 (2016)

\bibitem{CazenaveH15}
Cazenave, T., Hamida, S.B.: Forecasting financial volatility using nested monte
  carlo expression discovery. In: {IEEE} Symposium Series on Computational
  Intelligence, {SSCI} 2015, Cape Town, South Africa, December 7-10, 2015. pp.
  726--733 (2015). \doi{10.1109/SSCI.2015.110},
  \url{https://doi.org/10.1109/SSCI.2015.110}

\bibitem{CazenaveJ09}
Cazenave, T., Jouandeau, N.: Parallel nested monte-carlo search. In: 23rd
  {IEEE} International Symposium on Parallel and Distributed Processing,
  {IPDPS} 2009, Rome, Italy, May 23-29, 2009. pp.~1--6 (2009).
  \doi{10.1109/IPDPS.2009.5161122},
  \url{https://doi.org/10.1109/IPDPS.2009.5161122}

\bibitem{CazenaveSST16}
Cazenave, T., Saffidine, A., Schofield, M.J., Thielscher, M.: Nested monte
  carlo search for two-player games. In: Proceedings of the Thirtieth {AAAI}
  Conference on Artificial Intelligence, February 12-17, 2016, Phoenix,
  Arizona, {USA}. pp. 687--693 (2016),
  \url{http://www.aaai.org/ocs/index.php/AAAI/AAAI16/paper/view/12134}

\bibitem{cazenave2012tsptw}
Cazenave, T., Teytaud, F.: Application of the nested rollout policy adaptation
  algorithm to the traveling salesman problem with time windows. In: Learning
  and Intelligent Optimization - 6th International Conference, {LION} 6, Paris,
  France, January 16-20, 2012, Revised Selected Papers. pp. 42--54 (2012)

\bibitem{edelkamp2013algorithm}
Edelkamp, S., Gath, M., Cazenave, T., Teytaud, F.: Algorithm and knowledge
  engineering for the tsptw problem. In: Computational Intelligence in
  Scheduling (SCIS), 2013 IEEE Symposium on. pp. 44--51. IEEE (2013)

\bibitem{edelkamp2016monte}
Edelkamp, S., Gath, M., Greulich, C., Humann, M., Herzog, O., Lawo, M.:
  Monte-carlo tree search for logistics. In: Commercial Transport, pp.
  427--440. Springer International Publishing (2016)

\bibitem{edelkamp2014monte}
Edelkamp, S., Gath, M., Rohde, M.: Monte-carlo tree search for 3d packing with
  object orientation. In: KI 2014: Advances in Artificial Intelligence, pp.
  285--296. Springer International Publishing (2014)

\bibitem{edelkamp2014solving}
Edelkamp, S., Greulich, C.: Solving physical traveling salesman problems with
  policy adaptation. In: Computational Intelligence and Games (CIG), 2014 IEEE
  Conference on. pp.~1--8. IEEE (2014)

\bibitem{edelkamp2015monte}
Edelkamp, S., Tang, Z.: Monte-carlo tree search for the multiple sequence
  alignment problem. In: Eighth Annual Symposium on Combinatorial Search (2015)

\bibitem{Kocsis2006}
Kocsis, L., Szepesv\'ari, C.: Bandit based {M}onte-{C}arlo planning. In: 17th
  European Conference on Machine Learning (ECML'06). LNCS, vol.~4212, pp.
  282--293. Springer (2006)

\bibitem{koza1994genetic}
Koza, J.R., et~al.: Genetic programming II, vol.~17. MIT press Cambridge (1994)

\bibitem{langdon1997analysis}
Langdon, W.B., Poli, R., et~al.: An analysis of the max problem in genetic
  programming. Genetic Programming  \textbf{1}(997),  222--230 (1997)

\bibitem{Mehat2010}
M{\'e}hat, J., Cazenave, T.: Combining {UCT} and {Nested Monte Carlo Search}
  for single-player general game playing. {IEEE} Transactions on Computational
  Intelligence and {AI} in Games  \textbf{2}(4),  271--277 (2010)

\bibitem{Nagorko19}
Nagorko, A.: Parallel nested rollout policy adaptation. In: IEEE Conference on
  Games (2019)

\bibitem{NegrevergneC17}
N{\'{e}}grevergne, B., Cazenave, T.: Distributed nested rollout policy for
  samegame. In: Computer Games - 6th Workshop, {CGW} 2017, Held in Conjunction
  with the 26th International Conference on Artificial Intelligence, {IJCAI}
  2017, Melbourne, VIC, Australia, August, 20, 2017, Revised Selected Papers.
  pp. 108--120 (2017). \doi{10.1007/978-3-319-75931-9\_8}

\bibitem{portela2018unexpectedly}
Portela, F.: An unexpectedly effective monte carlo technique for the rna
  inverse folding problem. bioRxiv p. 345587 (2018)

\bibitem{potvin1996vehicle}
Potvin, J.Y., Bengio, S.: The vehicle routing problem with time windows part
  ii: genetic search. INFORMS journal on Computing  \textbf{8}(2),  165--172
  (1996)

\bibitem{PouldingF14}
Poulding, S.M., Feldt, R.: Generating structured test data with specific
  properties using nested monte-carlo search. In: Genetic and Evolutionary
  Computation Conference, {GECCO} '14, Vancouver, BC, Canada, July 12-16, 2014.
  pp. 1279--1286 (2014)

\bibitem{PouldingF15}
Poulding, S.M., Feldt, R.: Heuristic model checking using a monte-carlo tree
  search algorithm. In: Proceedings of the Genetic and Evolutionary Computation
  Conference, {GECCO} 2015, Madrid, Spain, July 11-15, 2015. pp. 1359--1366
  (2015)

\bibitem{RimmelEvo11}
Rimmel, A., Teytaud, F., Cazenave, T.: Optimization of the {Nested Monte-Carlo}
  algorithm on the traveling salesman problem with time windows. In:
  Applications of Evolutionary Computation - EvoApplications 2011: EvoCOMNET,
  EvoFIN, EvoHOT, EvoMUSART, EvoSTIM, and EvoTRANSLOG, Torino, Italy, April
  27-29, 2011, Proceedings, Part {II}. Lecture Notes in Computer Science,
  vol.~6625, pp. 501--510. Springer (2011)

\bibitem{Rosin2011}
Rosin, C.D.: Nested rollout policy adaptation for {Monte Carlo Tree Search}.
  In: IJCAI. pp. 649--654 (2011)

\end{thebibliography}

\end{document}